\documentclass[10 pt,journal]{IEEEtran}
\IEEEoverridecommandlockouts

\usepackage{fancyhdr}

\usepackage[normalem]{ulem}

\usepackage{
  tikz,
  pgfplots,
  pgfplotstable
}
\usepackage{hyperref}

\usetikzlibrary{
  shapes.geometric,
  arrows,
  external,
  pgfplots.groupplots,
  matrix
}

\pgfplotsset{compat=1.9}

\usepackage{mathtools,}

\DeclareMathAlphabet{\mathcal}{OMS}{cmsy}{m}{n}

\DeclareGraphicsExtensions{%
    .png,.PNG,%
    .pdf,.PDF,%
    .jpg,.mps,.jpeg,.jbig2,.jb2,.JPG,.JPEG,.JBIG2,.JB2}

\usepackage{gensymb}
\usepackage{cite}
\usepackage{amsmath,amssymb,amsfonts}
\usepackage{graphicx}
\usepackage{balance}
\usepackage{textcomp}
\usepackage{xcolor}

\def\BibTeX{{\rm B\kern-.05em{\sc i\kern-.025em b}\kern-.08em
    T\kern-.1667em\lower.7ex\hbox{E}\kern-.125emX}}
    
\usepackage{booktabs}
\usepackage{graphicx}
\usepackage{setspace}
\usepackage{caption}
\usepackage{comment}
\usepackage{enumitem}
\usepackage{multirow}
\usepackage{lipsum}
\usepackage{adjustbox}
\usepackage{algorithm}
\usepackage{algcompatible}
\usepackage{algpseudocode}
\usepackage{subcaption}
\usepackage{siunitx}
\usepackage{tabularx}
\usepackage{balance}
\usepackage[normalem]{ulem}
\usepackage{url}

\definecolor{black}{RGB}{0,0,0}
\definecolor{function}{RGB}{0,102,153}      
\definecolor{lightgreen}{RGB}{102,153,0}    
\definecolor{bluegreen}{RGB}{51,153,126}    
\definecolor{magenta}{RGB}{217,74,122}  
\definecolor{orange}{RGB}{226,102,26}       
\definecolor{purple}{RGB}{125,71,147}       
\definecolor{green}{RGB}{113,138,98}        

\begin{document}
\title{Enhancing Functional Safety in Automotive AMS Circuits through Unsupervised Machine Learning}

\makeatletter
\newcommand{\linebreakand}{%
  \end{@IEEEauthorhalign}
  \hfill\mbox{}\par
  \mbox{}\hfill\begin{@IEEEauthorhalign}
}
\makeatother

\author{\IEEEauthorblockN{Ayush Arunachalam\IEEEauthorrefmark{1},
Ian Kintz\IEEEauthorrefmark{1}, Suvadeep Banerjee\IEEEauthorrefmark{2}, Arnab Raha\IEEEauthorrefmark{2}, Xiankun Jin\IEEEauthorrefmark{3}, Fei Su\IEEEauthorrefmark{2}, \\
Viswanathan Pillai Prasanth\IEEEauthorrefmark{4}, Rubin A. Parekhji\IEEEauthorrefmark{4}, Suriyaprakash Natarajan\IEEEauthorrefmark{2}, and Kanad Basu\IEEEauthorrefmark{1}}

\IEEEauthorblockA{
\IEEEauthorrefmark{1}\textit{University of Texas at Dallas},
\IEEEauthorrefmark{2}\textit{Intel Corporation},
\IEEEauthorrefmark{3}\textit{NXP Semiconductors},
\IEEEauthorrefmark{4}\textit{Texas Instruments}}

\thanks {This work is supported in part by the Semiconductor Research Corporation (GRC Task: 2810.086). ({\textit{Corresponding Author}:\textit{ Ayush Arunachalam}}, Email: \textit{arunachalamayush@utdallas.edu}).

}
}

\date{}

\maketitle

\vspace{-0.140255in}

\begin{abstract}
Given the widespread use of safety-critical applications in the automotive field, it is crucial to ensure the Functional Safety (FuSa) of circuits and components within automotive systems. This includes electrical and electronic subsystems found in vehicles. \textcolor{black}{The Analog and Mixed-Signal (AMS) circuits prevalent in these systems are more vulnerable to faults induced by parametric perturbations, noise, environmental stress, and other factors, in comparison to their digital counterparts.} However, their continuous signal characteristics present an opportunity for early anomaly detection, enabling the implementation of safety mechanisms to prevent system failure. To address this need, we propose a novel framework based on unsupervised machine learning for early anomaly detection in AMS circuits. \textcolor{black}{The proposed approach involves injecting anomalies at various circuit locations and individual components to create a diverse and comprehensive anomaly dataset, followed by the extraction of features from the observed circuit signals. Subsequently, we employ clustering algorithms to facilitate anomaly detection. Specifically, we present a unique centroid selection technique that discerns the ideal cluster centroids to furnish optimum anomaly detection in AMS circuits. Finally, we propose a time series framework to enhance and expedite anomaly detection performance. Our approach encompasses a systematic analysis of anomaly abstraction at multiple levels pertaining to the automotive domain. Commencing at the hardware level, individual components and circuits are scrutinized to identify potential fault scenarios, focusing on AMS circuits. The analysis is then extended to a higher abstraction levels, culminating in an exploration of the impact of anomalies at the block-level.} \textcolor{black}{At each abstraction level, anomalies are injected to create diverse fault scenarios. By monitoring the system behavior under these anomalous conditions, we capture the propagation of anomalies and their effects at different abstraction levels, thereby potentially paving the way for the implementation of reliable safety mechanisms to ensure the FuSa of automotive SoCs.} The efficacy of the proposed solution was assessed using two AMS circuits (bandgap voltage reference and operational amplifier) frequently found in automotive systems. Our experimental findings indicate that our approach achieves 100\% anomaly detection accuracy and significantly optimizes the associated latency by 5$\times$, underscoring the effectiveness of our devised solution.

\end{abstract}


\fancyfoot[R]{\large Regular Paper}  

\section{Introduction} 
\label{sec:intro}

In recent times, there has been a significant rise in the use of Artificial Intelligence (AI) and autonomous driving in the automotive industry. The primary objective is to reduce human errors associated with vehicle operation. For instance, Advanced Driver Assistance Systems (ADAS) and autonomous driving enhance road safety and improve driving experiences. With the deployment of these technologies, ensuring the functional safety (FuSa) of modern automotive systems, which include various electrical and electronic (E/E) components, has become essential for reliable and secure vehicle operation.

Functional Safety, as defined in the International Organization for Standardization (ISO) 26262 standards~\cite{iso201126262}, refers to the absence of unreasonable risks caused by malfunctioning behaviors of E/E systems. These standards encompass a range of regulations, including, but not limited to validation, safety management, and risk analysis, specifically designed for E/E components in automotive systems. The presence of numerous semiconductor chips and Analog and Mixed-Signal (AMS) components in modern automotive systems makes them prone to faults, posing risks to the overall FuSa. To address this issue, we propose an anomaly detection approach focused on maintaining FuSa in automotive Systems-on-Chips (SoCs) that contain AMS components.

Faults in AMS circuits within automotive SoCs can compromise their FuSa and reliability. Several factors contribute to faults in these components~\cite{su2018improving}: (1) AMS components are more vulnerable to parametric perturbation and noise compared to their digital equivalents; (2) faults can arise due to environmental stressors such as heat, humidity, and vibration; (3) additionally, the complex interplay of system components, aging, latent defects, and single event upsets can manifest as faults in AMS components, potentially effectuating anomalous behavior. Such behavior, if undetected, may eventually lead to E/E system failure. Given that these automotive SoCs are deployed in critical and high-assurance environments, preserving and improving their FuSa during mission-critical operations is of utmost importance to avoid catastrophic situations and loss of lives.

\begin{figure}[t!]
\begin{subfigure}{0.5\columnwidth}
    \centering
    \includegraphics[width=\linewidth]{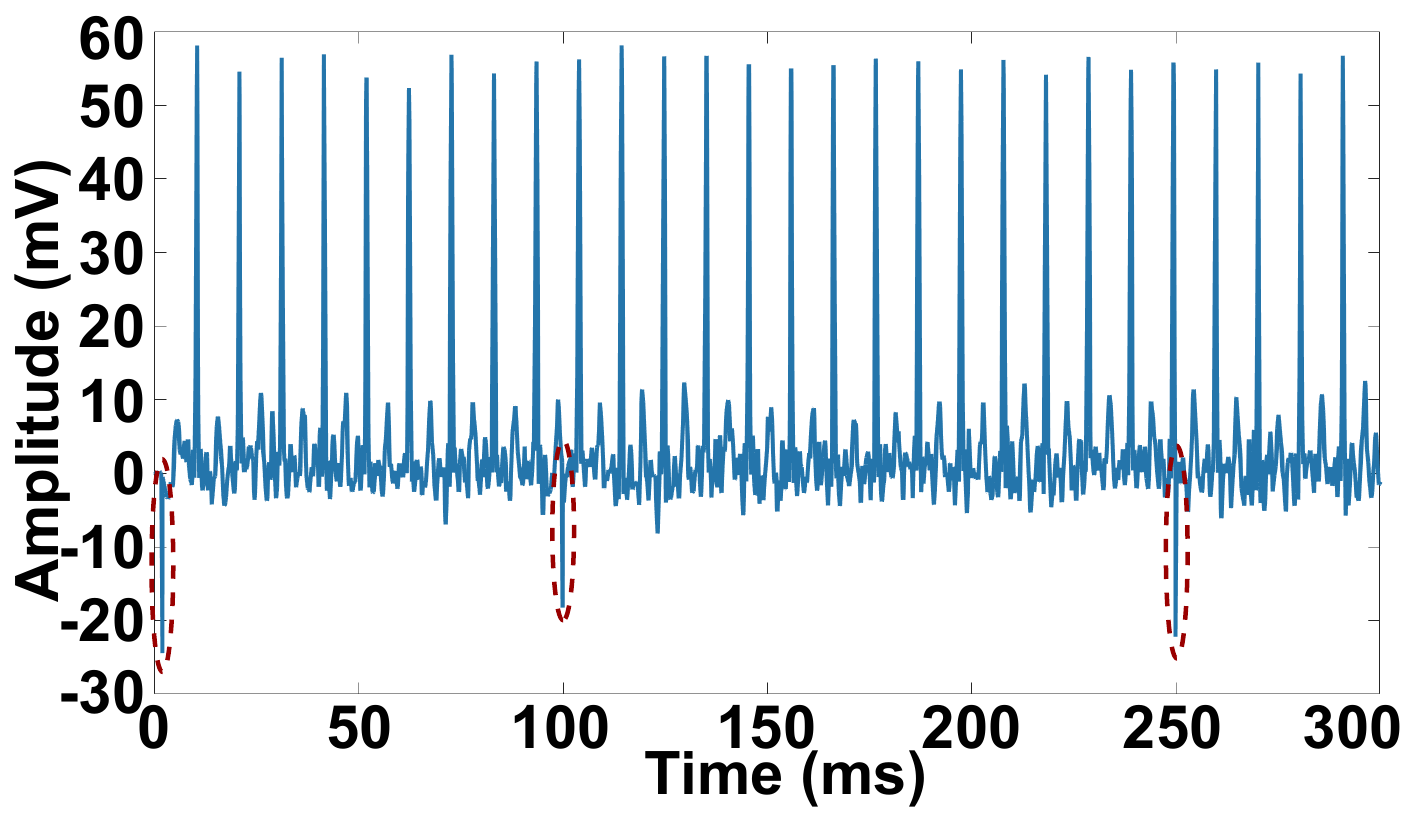}
    \caption{}
    \label{fig:pointanomaly}
\end{subfigure}%
~
\begin{subfigure}{0.5\columnwidth}
    \centering
    \includegraphics[width=\linewidth]{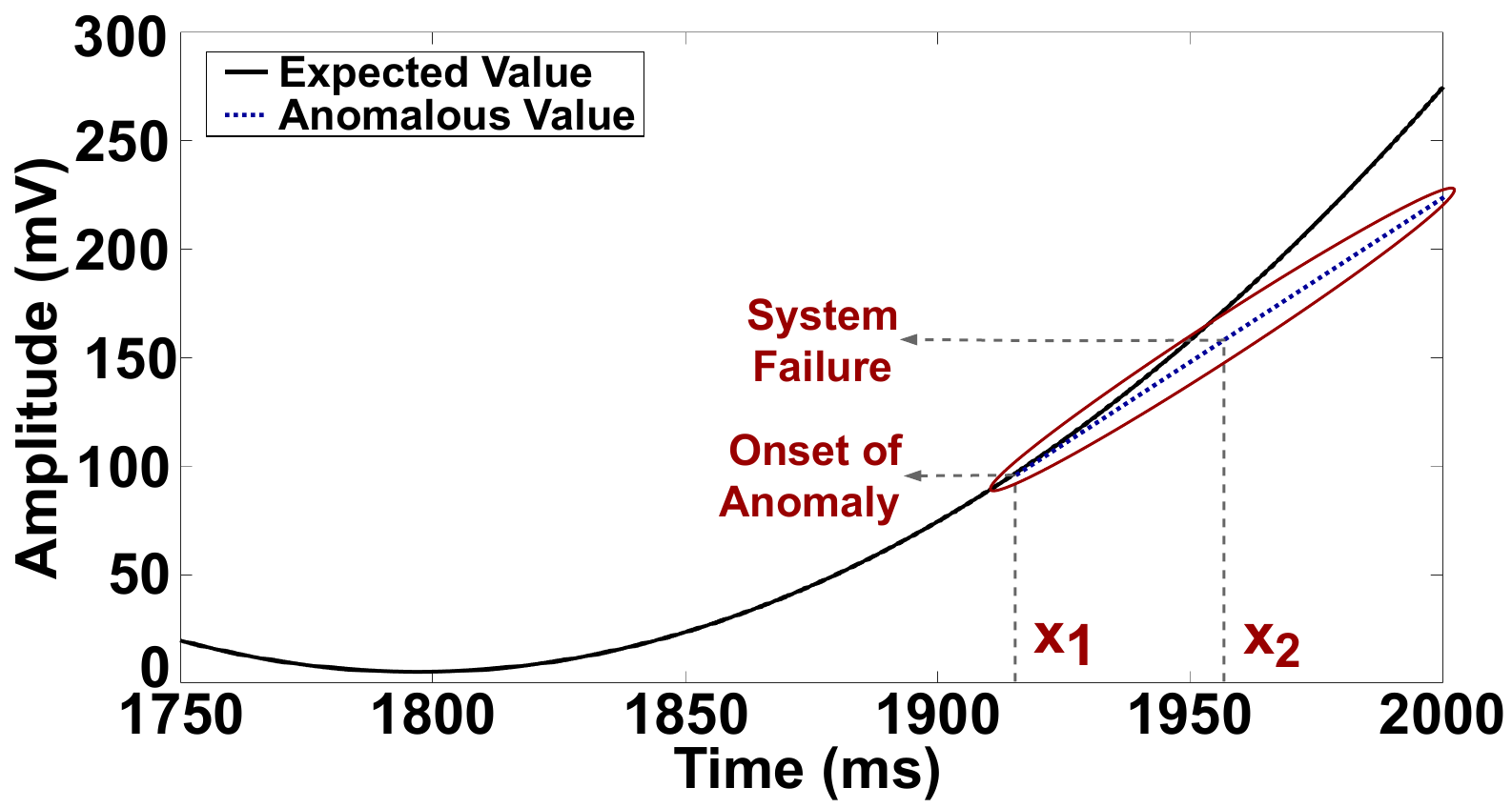}
    \caption{}
    \label{fig:trendanomaly}
\end{subfigure}
\vspace{0.1in}

\caption{Illustration of (a) point anomaly and (b) trend anomaly in a representative signal.}
\vspace{-0.1in}
\label{fig:anomalymotivation}
\end{figure}

Although automotive AMS components have certain limitations, they are inherently advantageous for analyzing the FuSa to enhance system reliability. AMS components operate in continuous signal regions, which can be leveraged for early anomaly detection. Figure~\ref{fig:anomalymotivation} illustrates different real-world anomalous scenarios encountered in automotive SoCs. Point anomalies, depicted in red in Figure~\ref{fig:pointanomaly}, represent discrete points that deviate from the rest of the signal. Trend anomalies, shown in red in Figure~\ref{fig:trendanomaly}, involve violations of the overall trend of signal data within a specific context. By analyzing the behavior of AMS components and detecting anomalies at an early stage, we can predict potential FuSa violations in automotive SoCs' AMS circuits. This proactive approach enables the implementation of safety mechanisms to prevent system failure. To accomplish this, we propose an unsupervised learning-based anomaly detection framework specifically designed for AMS components, offering robustness, flexibility, and suitability for real-world applications. We validate our approach using a Proof of Concept (PoC) involving two common automotive AMS circuits.

Anomalous behavior in Analog and Mixed-Signal (AMS) components lies in an intermediate state between safe operation and system failure. As AMS circuits often exhibit such anomalies prior to experiencing system failures, analyzing their behavior becomes crucial for predicting potential Functional Safety (FuSa) violations in automotive System-on-Chips (SoCs). By detecting anomalies at an early stage, proactive measures, such as implementing safety mechanisms, can be deployed to prevent system failures. \textcolor{black}{Moreover, anomaly abstraction plays a pivotal role in enhancing the understanding of system vulnerabilities and potential \textcolor{black}{safety hazards}. It involves a systematic analysis of fault scenarios at multiple abstraction levels, ranging from individual hardware components and circuits to higher block-level implementations. 
This paper presents a novel framework that utilizes unsupervised machine learning techniques for early anomaly detection in AMS circuits, aiming to propose a comprehensive solution for detecting FuSa violations at multiple hardware abstraction levels. The proposed anomaly abstraction methodology and subsequent analysis pave the way for the implementation of robust safety mechanisms and the assurance of functional safety in automotive SoCs, thereby fostering advancements in the domain of anomaly detection and safety assurance for safety-critical automotive applications. To this end, we propose an unsupervised learning-based anomaly detection framework specifically tailored for AMS components.} This framework is designed to be robust, flexible, and suitable for real-world applications. To validate the effectiveness of our approach, we perform a Proof of Concept (PoC) utilizing two AMS circuits frequently found in automotive SoCs, \emph{i.e.}, bandgap Voltage Reference (VRef) and Operational amplifier (Opamp) circuits. Through rigorous evaluation, we demonstrate the framework's capability to detect anomalies, thus providing valuable insights for enhancing safety measures in automotive systems. 

\textbf{Why unsupervised learning?} In this particular context, the choice of employing unsupervised learning can be justified by multiple compelling reasons: (1) Supervised learning relies on well-labeled datasets, which are challenging to obtain for AMS components of automotive SoCs. (2) Real-world scenarios may present a wide range of anomalies that are difficult to enumerate beforehand. Unsupervised learning provides flexibility in handling unforeseen scenarios encountered during operation. (3) By focusing on learning patterns from the dataset, unsupervised learning is advantageous when certain patterns are unknown to users and need to be identified. The key contributions of this paper are:

\begin{itemize}[itemsep=0pt,parsep=0pt,leftmargin=*]

    \item This article proposes an early anomaly detection framework, which, to the best of our knowledge, is the first unsupervised learning-based strategy to detect anomalies in automotive AMS circuits. 

    \item This article outlines an anomaly and fault injection framework, which encompasses various anomalous and fault scenarios, thereby creating an extensive anomaly dataset.

    \item This article develops a novel centroid selection algorithm that determines the optimum cluster centroids in order to furnish maximum anomaly detection performance.

    \item This article presents a time-series framework, which leverages the time-domain characteristics of AMS circuits, to accelerate anomaly detection, achieving a 5$\times$ reduction in associated latency.

    \item \textcolor{black}{This article investigates the impact of anomalies and faults on different levels of abstraction, thereby providing a comprehensive solution for the detection of automotive AMS FuSa violations.}

\end{itemize}

The subsequent sections of the paper are structured as follows: Section~\ref{sec:relatedwork} provides an overview of relevant previous studies in the field. Section~\ref{sec:methodology} elaborates on the proposed anomaly detection framework in detail. In Section~\ref{sec:exp_results}, the experimental results are presented and analyzed. Section~\ref{sec:discussion} offers a comprehensive discussion on the research, addressing frequently asked questions. Finally, Section~\ref{sec:conclusion} concludes the paper, offering a comprehensive summary of the obtained findings and results.
\section{Related Work}
\label{sec:relatedwork}

Over the past decade, researchers have focused on enhancing the functional safety of semiconductor devices, especially in the digital domain. Previous studies have predominantly emphasized on validating the FuSa of digital circuits in terms of logic and memory components. For instance, Error Correction Code (ECC), Logic Built-in Self-Test (LBIST), and in-field scan are few examples of techniques proposed to address this objective~\cite{tshagharyan2017effective}.

Anomaly detection, which involves identifying outliers or unusual patterns in data that do not conform to expected trends, has recently been integrated into the semiconductor industry to improve validation and testing quality and optimize manufacturing costs, specifically for AMS components ~\cite{krishnan2011robust, yilmaz2011adaptive}. The predominant focus of existing literature has been on detecting anomalies at the component level through the utilization of semi-supervised learning techniques~\cite{su2018improving}. This technique is associated with limitations, such as pre-specified feature space. These feature inputs to the anomaly detection model are typically predetermined and confined to a limited scale, rendering it inadequate for extensive feature spaces. \textcolor{black}{While existing techniques have primarily focused on FuSa validation at the component level, our proposed framework aims to enhance functional safety by addressing a higher abstraction level (block-level), in addition to the component level.} In this study, we present an unsupervised machine learning-based anomaly detection framework that specifically targets the improvement of automotive AMS FuSa. 
The key objective is to achieve early anomaly detection, providing a crucial opportunity for the implementation of proactive safety measures that can prevent system failures and ensure the reliable operation of automotive systems. \textcolor{black}{By addressing functional safety concerns at multiple abstraction levels, our proposed framework contributes to enhancing the overall safety and dependability of automotive systems.}

\begin{figure}[t!]
\centering
\includegraphics[width=0.9\linewidth]{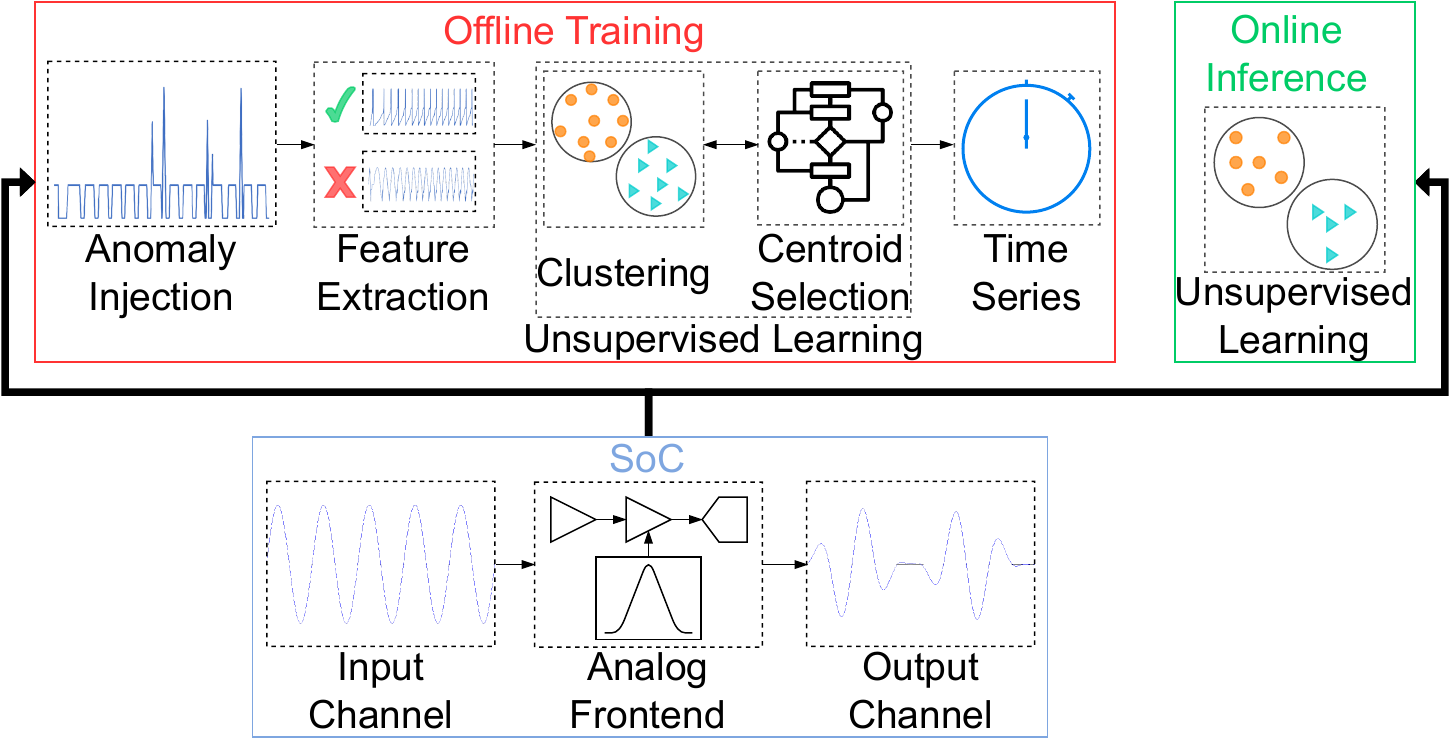}
\caption{Overview of the proposed anomaly detection framework for automotive AMS systems.}
\label{fig:detectionframework}
\vspace{-0.05in}
\end{figure}

\section{Proposed Methodology}
\label{sec:methodology}

Here, we describe the proposed methodology, which encompasses the following: (i) injection of diverse anomalies under various scenarios; (ii) feature extraction; (iii) unsupervised learning-based anomaly detection involving clustering and centroid selection, and (iv) time series framework. This approach is depicted by Figure~\ref{fig:detectionframework}. The initial stage of anomaly injection involves introducing anomalies in different components and locations of the circuits or case studies considered in this study. Subsequently, we generate datasets that comprise both anomalous and non-anomalous signals. Key features such as mean, variance, and slope are extracted from each type of signal to populate the dataset. Following this, the generated dataset is analyzed, revealing a Gaussian distribution of data associated with different anomalous scenarios, in the circuits considered in this study.
The subsequent stages entail the training of clustering algorithms, performing time series analysis, and concluding with inference.

\subsection{Anomaly Model}
\label{subsec:anomalyinjection}

This paper presents an anomaly model that incorporates the introduction of discrete point anomalies at diverse locations and within various components, covering a spectrum of anomalous scenarios. Point anomalies refer to individual points within a signal that deviate significantly from the rest~\cite{su2018improving}. In our study, these anomalies are injected into the output signals of different blocks within the circuit at discrete time instances. To illustrate our anomaly model, we utilize Figure~\ref{fig:voltagerefcircuit}, which depicts a bandgap voltage reference circuit consisting of four key blocks: Input (A), Phase-Locked Loop (PLL) (B), Trigonometric Function (Trig Fun) (C), and Output (D). Throughout the manuscript, Figure~\ref{fig:voltagerefcircuit} serves as an example circuit for illustrating our anomaly model. Anomalies are introduced at multiple block levels, notably within blocks A, B, and C, in our study. The injection rate of anomalies varies depending on the specific experiment and the signal being examined, with further details provided in Section~\ref{subsec:results}. Aside from injecting single-point anomalies, we also employ multipoint anomaly injection to evaluate how introducing anomalies simultaneously across multiple components affects the performance of the system. This investigation is crucial in shaping our anomaly detection approach, as it allows us to construct a more comprehensive detection framework by understanding the system behavior under diverse anomalous scenarios. For instance, multipoint anomaly injection involves the introduction of  anomalies at locations A and B, B and C, A and C, and A, B, and C simultaneously.

\begin{figure}[t!]
\centering
\includegraphics[width=0.9\linewidth]{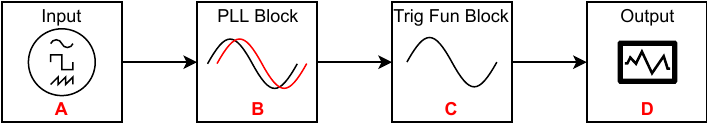}
\caption{Illustrative example circuit for anomaly model.}
 \label{fig:voltagerefcircuit}
\end{figure}

Subsequently, we conduct fault analysis at the component level. In this analysis, the circuits are represented using AMS components, such as Field-effect Transistors (FET), Bipolar Junction Transistors (BJT), resistors, capacitors, among others. We introduce anomalies in these components through the following methods: (1) Operating FETs in the linear/triode region instead of the saturation region, which corresponds to non-anomalous behavior~\cite{madian2010catastrophic}; (2) Parametric faults, which encompass scenarios where a particular circuit parameter is deliberately manipulated beyond its standard operational limits~\cite{guo2006coefficient, savir2003test}; (3) Open and short transistor faults~\cite{parky2009defect, arabi1997testing}. Open transistor faults are associated with the introduction of high impedance at either the source or gate terminal of FETs, whereas short transistor faults entail the use of low impedance models across the drain and source terminals of FETs. These various forms of anomalous situations facilitate the generation of a comprehensive dataset tailored for unsupervised learning algorithms, ensuring diversity in the data for effective model training. In our component-level fault analysis, we consider one output by observing the final circuit output. However, in the block-level analysis, in addition to the final block output, we also incorporate the output signals of intermediate blocks as observation signals.

\subsection{Unsupervised Learning Framework}
\label{subsec:anomalydetectionmethodology}

Following the injection of anomalies, datasets are carefully curated and compiled from the circuit to cover various anomalous scenarios. These datasets consist of both anomalous and non-anomalous data points, with equal representation to avoid introducing any bias. They serve as the basis for training and evaluating the unsupervised learning algorithms, which form the core of the proposed anomaly detection framework, as shown in Figure~\ref{fig:detectionframework}. Prior to applying these algorithms, relevant features such as mean, variance, and slope are extracted from the observation signal(s) comprising the generated datasets. For instance, an observation signal could be the final output signal of a circuit or an intermediate signal. In addition to these 1-dimensional features, we explore higher feature space dimensionality by utilizing multiple features. For instance, 2-dimensional tuples (mean and variance, variance and slope, mean and slope) and 3-dimensional tuples (all three features) contribute to our feature space. The clustering algorithms are trained and evaluated using this feature space. In this paper, we incorporate multiple clustering algorithms and analyze their performance in detecting anomalies in Section~\ref{sec:exp_results}. These algorithms aim to partition the data samples into two clusters, one representing anomalous signals and the other representing non-anomalous signals. The training phase is conducted offline, and the trained models are subsequently deployed. Finally, the inference phase, which involves anomaly detection during mission mode, is carried out online.

\subsection{Centroid Selection Algorithm}
\label{subsec:centroidalgo}

\begin{algorithm}[t!]
\begin{flushleft}
\textbf{Input}: $feature$, $\mu_k$, $\sigma_k$ \\

\textbf{Output}: $centroids$
\end{flushleft}

\caption{Centroid Selection}
\begin{algorithmic}[1]
    \State \textbf{Initialize} $centroids$ $\gets$ empty $Array$
    \State \textbf{Initialize} $k$ $\gets$ 2
    \State $\mu$ = \textit{mean}($feature$)
    \State $\sigma$ = \textit{std\_{dev}}($feature$)
    \For {j in $range(0, k)$}
    \If {$\mu_k[j]$ $<$ $\mu$}
        \State \textbf{Initialize} $varL$ $\gets$ empty $Array$
        \For {$i$ in $range(1, 5)$}
            \State $varL[i]$ $\gets$ abs($\mu_k[j]$+$i*\sigma_k[j]$-($\mu$-$i*\sigma$))
        \EndFor
    \EndIf
    \If{$\mu_k[j]$ $>$ $\mu$}
        \State \textbf{Initialize} $varG$ $\gets$ empty $Array$
        \For {$i$ in $range(1, 5)$}
            \State $varG[i]$ $\gets$ abs($\mu_k[j]$-$i*\sigma_k[j]$-($\mu$+$i*\sigma$))
        \EndFor
    \EndIf
    \EndFor
    \State $minValue \gets$ min($varL$)
    \For {$m$ in $varL$}
        \If {$minValue$ == $varL[m]$}
            \State $varL_k$ $\gets$ $m$
        \EndIf
    \EndFor
    \State \textbf{Repeat steps 20 to 24 to obtain $varG_k$ for $varG$}
    \State $centroids$ $\gets$ mean($feature[\mu_k[0]+(m+1)\sigma, \mu])$
    \State $centroids$ $\gets$ mean($feature[\mu, \mu_k[1]-(m+1)\sigma])$
    
\end{algorithmic}
\label{algo:centroidsel_algo}
\end{algorithm}

The unsupervised learning algorithms used in this study are generic and employed to identify clusters based on AMS data. However, these algorithms do not take into account the complex properties or relationships between the non-anomalous and anomalous data points in AMS circuits. As a result, their detection accuracy can be low in certain scenarios, as discussed in Section~\ref{subsec:results}. To address this limitation and optimize the detection accuracy of the proposed solution, we present a novel centroid selection algorithm (detailed in Algorithm~\ref{algo:centroidsel_algo}). To the best of our knowledge, this is the first approach specifically designed for AMS circuits. Our algorithm aims to identify cluster centroids that minimize the differences between the distributions over the entire sample space and the individual sub-sample spaces associated with the anomalous and non-anomalous AMS data points, thereby seeking to improve the fidelity of anomaly detection.

The inputs to Algorithm~\ref{algo:centroidsel_algo} comprise three main arguments: $feature$, $\mu_k$, and $\sigma_k$. The variable $feature$ denotes the normalized set of data samples, representing either the mean, variance, or slope. Meanwhile, $\mu_k$ and $\sigma_k$ are arrays containing mean and standard deviation values, respectively, obtained from the unsupervised learning algorithm for both non-anomalous ($\mu_k[0]$ and $\sigma_k[0]$) and anomalous ($\mu_k[1]$ and $\sigma_k[1]$) signals. The output of this algorithm is an array of length $k$, denoted as $centroids$, which contains the centroids for the desired number of clusters. In this particular case, since we are interested in identifying the anomalous and non-anomalous signals, $k$ is set to 2. Since the generated dataset exhibits a Gaussian distribution of data (as previously mentioned in Section~\ref{sec:methodology}), the algorithm begins by calculating the mean ($\mu$) and standard deviation ($\sigma$) of the entire sample space of $feature$ (lines 3 and 4), and subsequently compares $\mu_k$ with $\mu$ for different intervals of $\sigma_k$ and $\sigma$, as per Gaussian data distribution. For $\mu_k$ values less than $\mu$, the algorithm iterates through four intervals of $\sigma$ and $\sigma_k$ (line 8). At each interval, the absolute difference between $\mu_k$ + ($i*\sigma_k$) and $\mu$ - ($i*\sigma$) is computed and stored in an array called $varL$ (line 9). Similarly, for $\mu_k$ values greater than $\mu$, the algorithm iterates through the same four intervals (line 14), where the absolute difference between $\mu_k$ - ($i*\sigma_k$) and $\mu$ + ($i*\sigma$) is computed at each interval and stored in an array called $varG$ (line 15). Next, the algorithm identifies the indices corresponding to the minimum values in both $varL$ and $varG$ arrays (lines 19 and 20). These indices are subsequently employed to ascertain the optimal cluster centroids through the calculation of the average intersections between the Gaussian distribution of the $feature$ variable and both non-anomalous and anomalous signals (as indicated in lines 26 and 27).

\subsection{Time Series-based Analysis}
\label{subsec:timeseriesanalysis}

In this study, we propose a time series-based analysis to expedite anomaly detection in the AMS components of automotive SoCs, thereby enabling early anomaly detection. Such early detection is facilitated by the continuous signal regions that the AMS components operate in. Instead of considering the observation signal as a single entity of $N$ time instances, we employ a windowing approach to divide the signal into $k$ batches or intervals. Therefore, we now can extract $k$ sets of features, as opposed to a single set. Each window now covers $N/k$ time instances, resulting in $k$ sets of features ($k \in \mathbb{Z}$). Consequently, a new dataset is generated, consisting of $k$ sets of mean, variance, and slope features for both anomalous and non-anomalous signals. This dataset is subsequently used for training and evaluating various unsupervised learning-based clustering algorithms.

\begin{figure}[t!]
\begin{subfigure}{0.5\columnwidth}
    \centering
    \includegraphics[width=\linewidth]{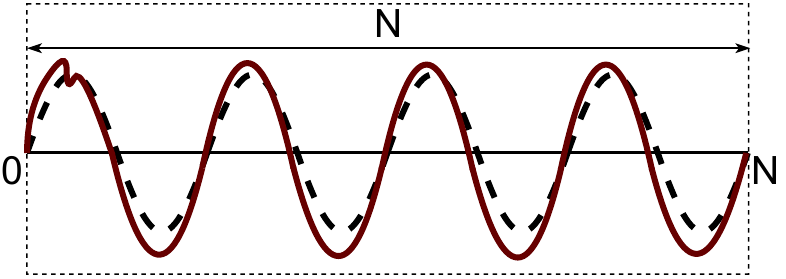}
    \caption{}
    \label{fig:nontimeseries_singlewindow}
\end{subfigure}%
~
\begin{subfigure}{0.5\columnwidth}
    \centering
    \includegraphics[width=\linewidth]{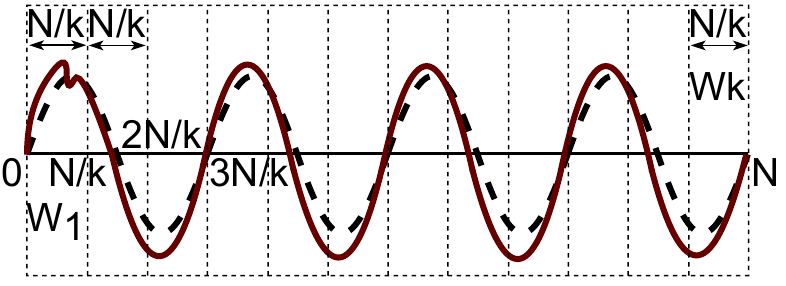}
    \caption{}
    \label{fig:timeseries_nwindows}
\end{subfigure}
\vspace{0.1in}
\caption{Illustrative example of (a) non time series approach and (b) time series approach.
}
\label{fig:timeseries_earlyanomdetection}
\vspace{-0.1in}
\end{figure}

Figure~\ref{fig:timeseries_earlyanomdetection} provides an illustrative example that highlights the advantages of utilizing time series-based analysis for early anomaly detection. As seen in Figure~\ref{fig:timeseries_earlyanomdetection} the non-anomalous signal and its anomalous counterpart are highlighted in black and red, respectively. Figure~\ref{fig:nontimeseries_singlewindow} illustrates the non time series-based technique, while Figure~\ref{fig:timeseries_nwindows} showcases the proposed time series approach in our study. In the non time series-based technique, a single window covering the entire duration of the signal (i.e., $N$ time units) is considered, resulting in a single set of features being extracted from both the non-anomalous and anomalous signals. This implies that the complete time period or length of the signals must be observed to identify potential anomalous behavior. In contrast, the time series-based approach incorporates $k$ windows, each with a size of $N/k$ time units. However, it is not necessary to consider all $k$ windows (and their corresponding sets of features) to detect anomalies. Instead, we propose detecting anomalies using $m$ sets of features, where $m<k$. This method reduces the detection time to $m*N/k$ time units, eliminating the need to wait for the entire $N$ time units before performing anomaly detection. As a result, this approach offers the advantage of early anomaly detection, enabling us to identify anomalies before they lead to eventual system failure.

\subsection{Anomaly Abstraction}
\label{subsec:anomalyabstraction}

\textcolor{black}{In this research, we adopt a systematic methodology to perform anomaly abstraction, wherein we consider an operational amplifier circuit (component-level) as the base symbol. To create a comprehensive anomaly model, we construct k-stage non-inverting amplifiers by iteratively replicating and interconnecting the base symbol. Each k-stage amplifier embodies a specific abstraction of the Opamp circuit, encapsulating its functional characteristics and interrelationships, specifically at the block-level. To develop an effective anomaly model, we carefully study the behavior of the k-stage amplifier under various normal operating conditions. Subsequently, we introduce anomalous variations that encompass potential deviations from the normal behavior, as will be described in Section~\ref{subsubsec:anomalyabstractionresults}. These anomalies are modeled to capture a wide range of possible fault scenarios that the Opamp circuit might encounter during its operation. Furthermore, to explore the impact of anomalies in different configurations, we consider diverse anomalous combinations by injecting the anomalous symbol at various locations within the k-stage amplifier. Figure~\ref{fig:threestageamp} illustrates an example one such amplifier, called Tri-stage amplifier (where k = 3). This approach allows us to investigate the effects of anomalies at different stages and understand how such deviations propagate through the system. By analyzing multiple scenarios of anomaly injection, we gain valuable insights into the response of the k-stage amplifier and its vulnerability to faults. This in-depth examination facilitates a comprehensive analysis of the non-inverting amplifier behavior under different anomalous situations and paves the way for effective anomaly detection and mitigation strategies at a higher abstraction level.}

\begin{figure}[b!]
\centering
\includegraphics[width=0.9\linewidth]{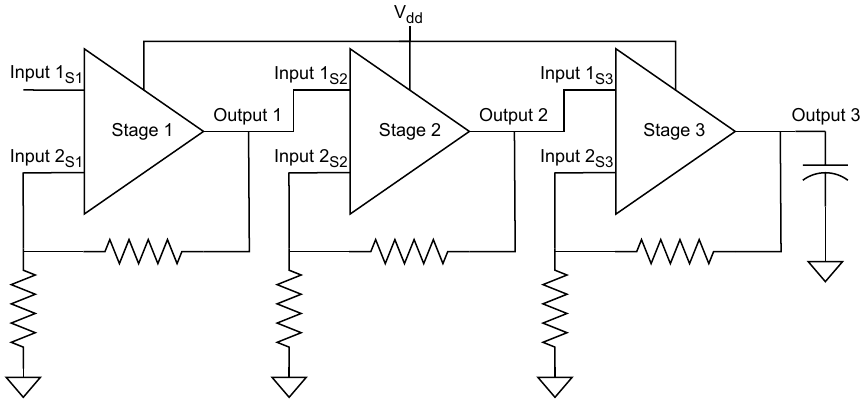}
\caption{Illustration of a Tri-stage amplifier, where k = 3.} 
\label{fig:threestageamp}
\end{figure}

\begin{algorithm}[t!]
\begin{flushleft}
    \Statex \textbf{Input:} cktSpecs, gain, anomList
    \Statex \textbf{Output:} kStAmp, kStAmpAN
\end{flushleft}
  \caption{Anomaly Abstraction}
  \begin{algorithmic}[1]

    \Function{CktSym}{cktSpecs, gain}
    \State $ckt \gets Design(cktSpecs, gain)$
    \State $ckt \gets Simulate(ckt)$
    \State \Return $cktSym \gets Symbol(ckt)$
    \EndFunction

    \Function{InjAnom}{ckt, anomList}
    \State $cktAN \gets InjAnom(ckt, anomList)$
    \State \Return $cktSymAN \gets Symbol(cktAN)$    
    \EndFunction

    \Function{DesAmp}{cktSym}
    \State Define $gainAmp$
    \State $kStAmp \gets Design(cktSym)$
    \State \Return $kStAmp \gets Simulate(kStAmp)$ 
    \EndFunction

    \Function{DesAmpAN}{cktSymAN}
    \State $kStAmpAN \gets Design(cktSymAN)$
    \State \Return $kStAmpAN \gets Simulate(kStAmpAN)$ 
    \EndFunction

    \State \textbf{Main:}
    \State \Call{CktSym}{cktSpecs, gain}
    \State \Call{InjAnom}{ckt, anomList}
    \State \Call{DesAmp}{cktSym}
    \State \Call{DesAmpAN}{cktSymAN}
    \State \textbf{Output:} kStAmp, kStAmpAN
  \end{algorithmic}
\label{algo:anomalyabstraction}
\end{algorithm}

\textcolor{black}{Algorithm~\ref{algo:anomalyabstraction} describes the proposed anomaly abstraction technique. Although we outline a generic methodology to design block-level implementations of amplifiers, we focus on k-stage non-inverting amplifiers constructed using Opamp circuits. In this regard, we have explained the application of Algorithm~\ref{algo:anomalyabstraction} within this context. The inputs to Algorithm~\ref{algo:anomalyabstraction} are circuit specifications ($cktSpecs$), desired gain ($gain$), and anomaly list ($anomList$), while the outputs are the non-anomalous and anomalous versions of the desired k-stage non-inverting amplifier ($kStAmp$ and $kStAmpAN$, respectively). As outlined in Algorithm~\ref{algo:anomalyabstraction}, the first step involves the design of the operational amplifier circuit according to the user-defined specifications (outlined in the AMS benchmark suite~\cite{sunter2017ms}) and the generation of a corresponding symbol. This step, as defined in the $CktSym$ function, is associated with the design of a non-anomalous Opamp circuit ($ckt$), which is simulated and tested for functionality and performance, as per the desired requirements.
Following this, we inject anomalies in the circuit using the specified anomalies, \emph{i.e.}, operating mode anomalies pertaining to PFETs and NFETs, and generate its corresponding symbol, as defined in the $InjAnom$ function. Following this, we design the non-anomalous ($kStAmp$) and anomalous ($kStAmpAN$) k-stage non-inverting amplifier, as defined in $DesAmp$ and $DesAmpAN$ functions, respectively. $kStAmp$ is designed according to user-defined gain ($gainAmp$), similar to $ckt$. Unlike $kStAmp$, which has a single design, $kStAmpAN$, has multiple configurations, based on the type of anomaly injected and location of the anomalous symbol.
}
\section{Experimental Results}
\label{sec:exp_results}

\subsection{Experimental Setup}
\label{subsec:setup}

In this research paper, we employ a Proof-of-Concept (PoC) approach consisting of two case studies to assess the proposed anomaly detection framework. The first case study involves a VRef circuit, as depicted in Figure ~\ref{fig:voltagerefcircuit}, which serves as an illustrative example. The second case study focuses on an Opamp circuit. Both case studies exemplify Analog and Mixed-Signal (AMS) circuits commonly encountered in automotive System-on-Chips (SoCs)~\cite{sunter2017ms}. This paper explores four distinct clustering algorithms for our evaluation. Firstly, we consider the k-means algorithm, a centroid-based hard clustering method that groups data samples into \textit{k} clusters based on their Euclidean distance from cluster centroids or mean values~\cite{lloyd1982least, hartigan1979algorithm, likas2003global}. Secondly, we examine the Gaussian Mixture Model (GMM), a probabilistic model that facilitates soft clustering by assigning probabilities to each data point, indicating their likelihood of belonging to different clusters. GMM utilizes Gaussian density functions for this purpose~\cite{reynolds2009gaussian, rasmussen1999infinite, mclachlan2014number}. Thirdly, we analyze the Balanced Iterative Reducing and Clustering using Hierarchies (BIRCH) algorithm, a hierarchical clustering approach designed to handle noise or outliers in datasets. BIRCH demonstrates scalability with high-dimensional data, which is particularly advantageous in our context~\cite{zhang1996birch, zhang1997birch, lorbeer2018variations}. Finally, we investigate Spectral clustering, which employs eigenvalues of similarity matrices to reduce the dimensionality of the data and subsequently perform clustering in a lower-dimensional space~\cite{ng2002spectral, bach2003learning, von2007tutorial}. We evaluate the performance of these clustering algorithms based on their accuracy in detecting anomalies. Specifically, we measure the correctness of assigning data points to clusters representing anomalous and non-anomalous signals. This assessment serves as a key indicator of the effectiveness of the anomaly detection framework introduced in this study.

\subsection{Results}
\label{subsec:results}

\subsubsection{Voltage Reference Circuit: Block-level Analysis}
\label{subsubsec:vrefresults}

In the initial phase of our investigation, we focus on evaluating the accuracy of anomaly detection pertaining to anomalies injected into the input and intermediate circuit components of the VRef circuit, as depicted in Figure~\ref{fig:voltagerefcircuit}. To assess the performance of the proposed anomaly detection framework comprehensively, we conduct experiments involving various types of anomalies. Specifically, we consider Input Anomaly (IA), PLL block Anomaly (PA), and Trig Fun block Anomaly (TA) experiments. Subsequently, we proceed with experiments involving periodic anomalies injected into the circuit. These experiments encompass combinations such as Input-PLL block Periodic Anomaly (IPPA), Input-Trig Fun block Periodic Anomaly (ITPA), PLL-Trig Fun block Periodic Anomaly (PTPA), and Input-PLL-Trig Fun block Periodic Anomaly (IPTPA). Furthermore, we explore experiments involving the injection of random anomalies. This set of experiments includes scenarios like Input-PLL block Random Anomaly (IPRA), Input-Trig Fun block Random Anomaly (ITRA), PLL-Trig Fun block Random Anomaly (PTRA), and Input-PLL-Trig Fun block Random Anomaly (IPTRA). Additionally, for the analysis of multipoint signal observations, as elucidated in subsequent sections, we incorporate experiments specifically focusing on PLL block Periodic Anomaly (PPA) and PLL block Random Anomaly (PRA), in conjunction with the aforementioned experiments. It is pertinent to mention that the examination of multiple signals necessitates the exclusion of anomalies solely in the Trig Fun block, which precedes only one block (D), in contrast to the input and PLL blocks (which precede three and two blocks, respectively), as illustrated in Figure~\ref{fig:voltagerefcircuit}. The magnitude or strength of the anomalies introduced into the different blocks is contingent upon the signals that form them. These `delta values' are calculated based on the signal intensity observed during non-anomalous operation. This approach ensures that the anomalies accurately reflect the deviations from normal behavior and maintain consistency with the signal characteristics under normal operating conditions.

\paragraph{Single-point Anomaly Injection}

In this experiment, we introduce point anomalies into the time domain of the input signal with amplitudes ranging from $2\times$ to $5\times$ of the maximum signal strength. The injection of these anomalies follows a random pattern, where they are inserted into the signal at various time instances. The rate of anomaly injection is defined as the percentage of anomalies being introduced relative to the signal length, which is determined by the discrete time instances constituting the signal. We vary this rate between 0.1\% and 0.5\% of the signal length, with increments of 0.1\% for each step. This allows us to systematically assess the impact of different anomaly densities on the anomaly detection performance of the proposed framework. In contrast, the injection of point anomalies in intermediate components (PLL and Trig Fun blocks) follows a periodic pattern, where the anomalies are inserted at user-specified time instances corresponding to 90\% of the maximum signal intensity. To achieve this, the amplitude is increased by a value equivalent to the maximum signal intensity at the specified time instances. Furthermore, in addition to the periodic anomalies, we investigate the effect of injecting random anomalies into the intermediate components. This will allow us to comprehensively evaluate the anomaly detection capabilities of our proposed framework under different anomalous scenarios. The outcomes of these experiments are depicted in Figure~\ref{fig:table1results}.

\begin{figure}[t!]
\centering
\includegraphics[width=0.8\linewidth]{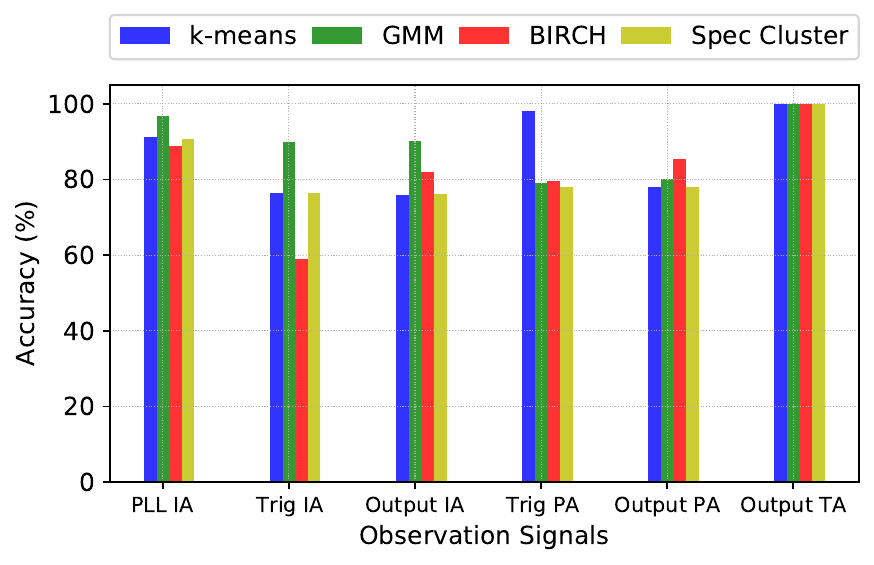}
\caption{Evaluation of unsupervised learning models for the injection of anomalies introduced in a random manner at the input and periodic manner in intermediate VRef components.}
\label{fig:table1results}
\vspace{-0.125in}
\end{figure}

As demonstrated in Figure~\ref{fig:table1results}, GMM exhibits the highest detection accuracy across the majority of our experiments, achieving up to 96.5\% accuracy when considering all features of the \textit{PLL} signal in the IA experiment. This remarkable performance is attributed to the aggregated-feature space utilized, where the mean, variance, and slope are represented as a single 3-dimensional tuple. The effectiveness of GMM is particularly prominent in scenarios where the data exhibits a Gaussian distribution, which is prevalent in our datasets. Furthermore, we have observed that the detection performance is dependent on the specific signal being monitored. Based on our results, it can be inferred that our anomaly detection framework furnishes the best performance by monitoring the output signal of the block immediately following the anomalous block. This observation highlights the importance of considering the behavior of the subsequent block, since it tends to be more indicative of the impact and manifestation of the injected anomaly. For instance, in the TA experiment, monitoring the \textit{Output} signal results in an improved performance of 100\% detection accuracy, compared to the PA experiment, which achieves 78\% detection accuracy. This observation is significant as the succeeding block experiences the maximum impact of the injected anomaly, making its output signal an optimal choice for achieving the best anomaly detection performance.

\paragraph{Multipoint Anomaly Injection}

This section involves the injection of point anomalies simultaneously into multiple components. Both periodic and random anomalies are considered in this evaluation. For the input anomaly injection, point anomalies are randomly inserted in the time domain, varying between 0.1\% to 0.5\% of the signal length. On the other hand, anomaly injection in the intermediate components follows a periodic pattern. We insert point anomalies into the time domain at various user-specified thresholds. Specifically, we employ two thresholds, \emph{i.e.}, 10\% and 90\% of the maximum signal intensity. These thresholds are used to insert anomalies of different intensities, corresponding to 5\% and 10\% of the maximum signal intensity. In this set of experiments, the magnitude of the inserted anomalies is comparatively modest when contrasted with the earlier experiments, signifying a moderate difference in signal intensity at the anomalous locations. The results of this evaluation are depicted in Figure~\ref{fig:table3aacc}. It is evident from the results that GMM outperforms other algorithms in the majority of our experiments. Specifically, GMM achieves up to 100\% anomaly detection accuracy by observing the mean of the \textit{Output} signal in the ITPA, PTPA, and IPTPA experiments.

\begin{figure}[t!]
\begin{subfigure}{0.5\columnwidth}
    \centering
    \includegraphics[width=\linewidth]{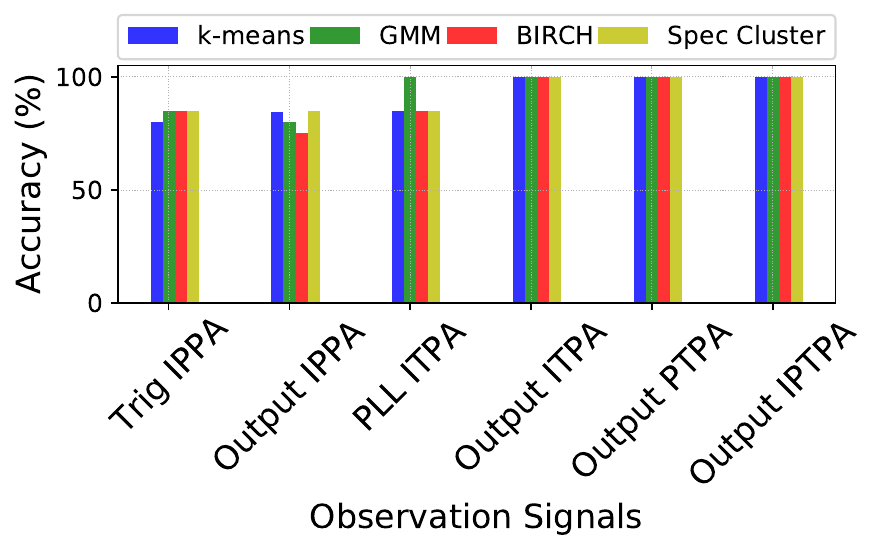}
    \caption{}
    \label{fig:table3aacc}
\end{subfigure}%
~
\begin{subfigure}{0.5\columnwidth}
    \centering
    \includegraphics[width=\linewidth]{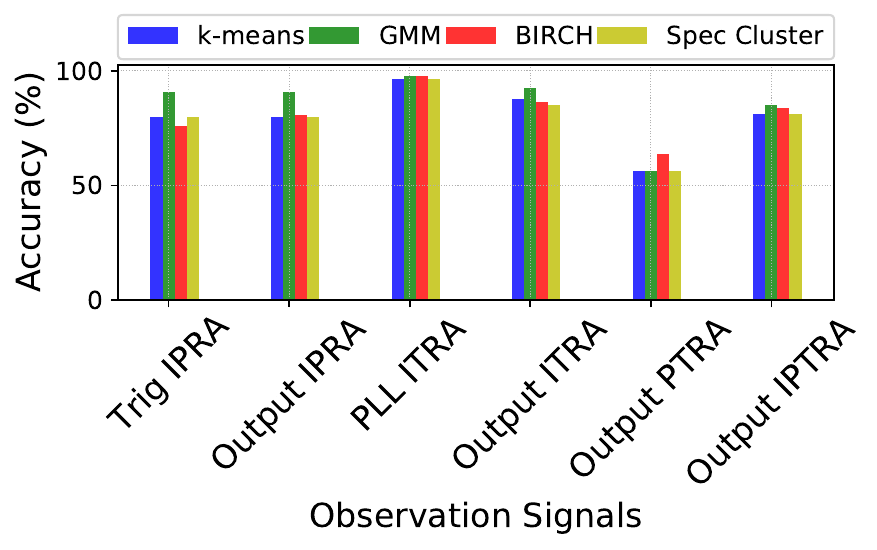}
    \caption{}
    \label{fig:table3bacc}
\end{subfigure}
\vspace{0.05in}
\caption{Evaluation of unsupervised learning models for (a) periodic and (b) random multipoint anomalies in VRef circuit.}
\label{fig:table3results}
\vspace{-0.05in}
\end{figure}

In the second set of experiments, we introduce random point anomalies with an injection rate varying between 0.1\% to 0.5\% (in increments of 0.1\%) in all experiments. The results of these experiments are presented in Figure~\ref{fig:table3bacc}, with the X-axis representing the signals observed in each experiment. From Figure~\ref{fig:table3bacc}, it is evident that GMM consistently outperforms other methods in the majority of our experiments. Notably, GMM achieves up to 97.5\% accuracy by monitoring the \textit{PLL} signal using the aggregated-feature space (ITRA experiment). 

In both sets of experiments, GMM consistently exhibits the best detection performance. The primary factor responsible for this can be traced back to the datasets generated, which, as observed in Section~\ref{subsubsec:vrefresults}a, also follow a Gaussian distribution of data. The suitability of GMM for Gaussian data contributes to its superior performance in accurately detecting anomalies. Furthermore, in accordance with the results outlined in Section~\ref{subsubsec:vrefresults}a, the optimal detection performance is consistently obtained by monitoring the output signal of the block situated directly after the anomalous block. This observation, as explained in Section~\ref{subsubsec:vrefresults}a, highlights the significance of monitoring the subsequent block's behavior to effectively detect anomalies in our proposed framework.

\paragraph{Multipoint Signal Observation}
Based on the consistently superior detection performance achieved by the aggregated-feature approach in the majority of our experiments, we now aim to explore further increased feature space dimensionality. This involves observing features of multiple signals simultaneously, resulting in $N$-dimensional tuples, where $N$ is determined by the number of signals and features being monitored. In this exploration, we conduct experiments involving the injection of point anomalies in both periodic and random manners. For random anomaly insertion, we vary the injection rates between 0.1\% to 0.5\%. On the other hand, for periodic anomaly injection, we employ multiple thresholds, as specified earlier in Section~\ref{subsubsec:vrefresults}. These thresholds correspond to different percentages of the maximum signal intensity. The results of this set of experiments are presented in Figure~\ref{fig:table4acc}. By observing the performance of the proposed framework in this extended feature space dimensionality, we aim to gain insights into the impact of monitoring multiple signals and their features concurrently on anomaly detection accuracy.

As delineated in Figure~\ref{fig:table4acc}, our empirical findings consistently underscore the superior anomaly detection performance of GMM. We achieve accuracy rates of 100\% across a spectrum of experiments, namely IA, ITPA, and ITRA, when scrutinizing multidimensional feature tuples. For instance, in the IA experiment, a 2-dimensional tuple comprising the mean values of \textit{PLL} and \textit{Trig} signals results in the best detection accuracy. In the ITPA and ITRA experiments, 3-dimensional tuples consisting of the mean values of the \textit{Output} signal and the two \textit{PLL} components, namely frequency and intensity, yield the highest accuracy. The effectiveness of GMM can be attributed to the same factors observed in Section~\ref{subsubsec:vrefresults}a and Section~\ref{subsubsec:vrefresults}b. It is particularly successful in capturing the Gaussian distribution of data in the feature spaces utilized in our experiments. Furthermore, the feature spaces that furnish the maximum detection accuracy primarily involve features of the signals that are directly downstream of the anomalous block, aligning with the observations in Section~\ref{subsubsec:vrefresults}a and Section~\ref{subsubsec:vrefresults}b. This underscores the importance of monitoring the behavior of the subsequent blocks to achieve enhanced anomaly detection. Importantly, the experimental results of these multipoint signal observation experiments demonstrate significantly higher anomaly detection performance compared to Section~\ref{subsubsec:vrefresults}a and Section~\ref{subsubsec:vrefresults}b, resulting in accuracy improvements ranging from 2.5\% to 8.85\%. This underscores the advantages of multipoint observation, where the concurrent observation of multiple features and signals yields more profound insights, ultimately resulting in enhanced detection performance.

\begin{figure}[t!]
\begin{subfigure}{0.5\columnwidth}
    \centering
    \includegraphics[width=\linewidth]{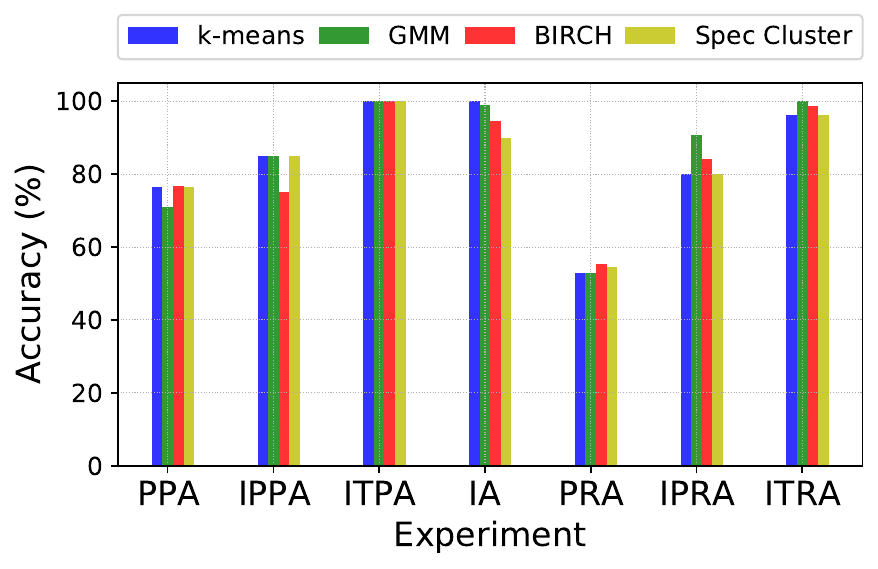}
    \caption{}
    \label{fig:table4acc}
\end{subfigure}%
~
\begin{subfigure}{0.5\columnwidth}
    \centering
    \includegraphics[width=\linewidth]{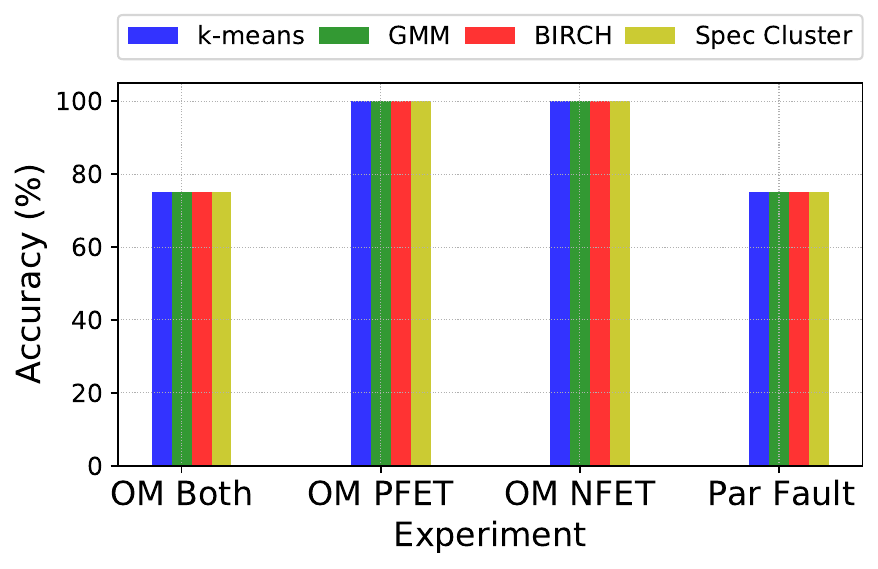}
    \caption{}
    \label{fig:table6acc}
\end{subfigure}
\vspace{0.1in}
\caption{Performance of unsupervised learning models (a) via multipoint observation for periodic and random anomalies and (b) for operating mode anomalies pertaining to FETs in bandgap voltage reference circuit.
}
\label{fig:table6results}
\vspace{-0.15in}
\end{figure}

\subsubsection{Voltage Reference Circuit: Component-level Analysis}
\label{subsubsec:vref_spiceresults}

In this set of experiments, we expand our analysis to explore the injection of anomalies into specific circuit components within the VRef circuit, as depicted in Figure~\ref{fig:circuits_spice}a. Specifically, we investigate the anomalous behavior of Field-effect Transistors (FETs), distinguished by their functioning within the triode/linear operational region, which represents a departure from their standard operation within the saturation region. Additionally, we introduce parametric faults in the VRef circuit, focusing on deviations related to the operating temperature. Anomalous behavior in this context refers to the operation of FETs outside the specified temperature range of -40$\degree$C to 125$\degree$C. Consequently, we conduct four distinct experiments as follows: (i) Anomalous Operating Mode of both PFETs and NFETs in the circuit (OM Both); (ii) Anomalous Operating Mode of only PFETs (OM PFET); (iii) Anomalous Operating Mode of only NFETs (OM NFET); (iv) Parametric fault (Par Fault). The results of these experiments are presented in Figure~\ref{fig:table6acc}, where the X-axis corresponds to the various experiments. This analysis allows us to gain insights into the detection performance of the proposed framework concerning specific anomalous behaviors in individual circuit components and parametric faults in the VRef circuit. In this experiment, we analyze the circuit using transient analysis, where the output voltage is varied with time, and two variants of DC analysis, where the output voltage is varied input voltage, and operating temperature, respectively.

\paragraph{FET Operating Mode Anomalies and Parametric Faults}
Figure~\ref{fig:table6acc} indicates consistent anomaly detection performance across all algorithms, achieving up to 100\% accuracy by monitoring the mean value of the circuit output in both OM PFET and OM NFET experiments, which demonstrates the effectiveness of the proposed solution. The results remain consistent across all three types of analyses. The uniformity in detection performance can be attributed to the substantial impact of these anomalies and faults on the circuit output, leading to distinguishable deviations from the normal, non-anomalous output. The mean feature, capturing changes in signal intensity or amplitude, proves to be particularly effective in detecting and identifying these deviations, thus yielding maximum detection accuracy.

\paragraph{FET Open and Short Faults}
In this experiment, we inject open and short transistor faults in the VRef circuit. Transistor short faults are associated with the shorting of Drain-Source terminals of an FET using a \SI{100}{\ohm} resistor. On the other hand, open faults are simulated by inserting a high resistance of \SI{1}{\giga\ohm} in the source terminal of FETs. To identify the transistor that demonstrates the maximum impact in terms of output deviation from its normal, non-anomalous counterpart, we conducted a comprehensive analysis. Our findings indicate that a fault in the FET located closest to the output section, as depicted in Figure~\ref{fig:circuits_spice}a, induces the most significant impact on the output of the VRef circuit. 

\begin{table}[b!]
\caption{Evaluation of the unsupervised learning-based approach for transistor faults in VRef circuit.}
\footnotesize
\centering
\begin{adjustbox}{width=0.95\columnwidth}
\begin{tabular}{|c|c|c|c|c|}
\hline
\multicolumn{1}{|c|}{\multirow{2}{*}{\textbf{Fault}}} & \multicolumn{4}{c|}{\textbf{Accuracy (\%)}}    \\ \cline{2-5}
\multicolumn{1}{|c|}{}    & \textbf{k-means} & \textbf{GMM} & \textbf{BIRCH} & \textbf{Spec Cluster} \\ \hline \hline
Open    & 81      & 85  & 70    & 75    \\  \hline
Short   & 85      & 88  & 70    & 75    \\ \hline

\end{tabular}
\end{adjustbox}
\label{table:kNNresults_spice_bgr_transistorfault}
\end{table}

\begin{figure}[t!]
\centering
\includegraphics[width=\linewidth]{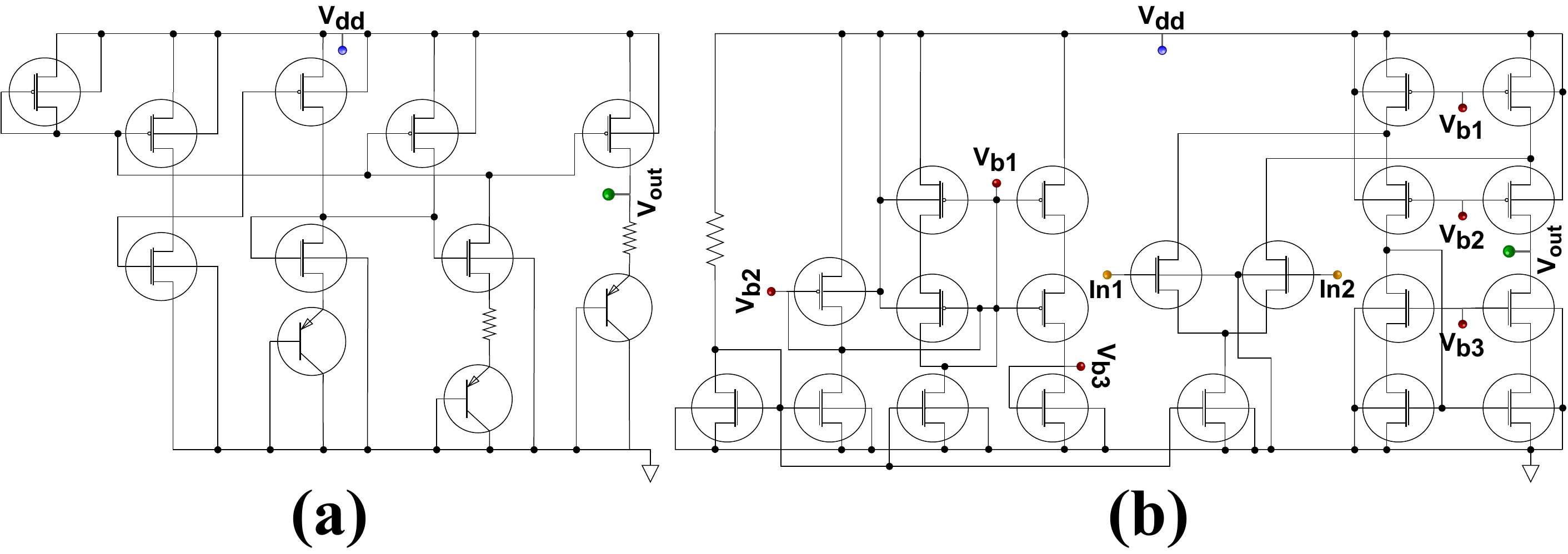}
\caption{Depiction of (a) bandgap voltage reference and (b) operational amplifier circuits for component-level analysis.}
 \label{fig:circuits_spice}
\end{figure}

As seen in Table~\ref{table:kNNresults_spice_bgr_transistorfault}, GMM outperforms the other algorithms, producing 85\% detection accuracy for transistor open faults and 88\% for transistor short faults, utilizing the mean feature. GMM's superior performance can be attributed to the Gaussian distribution of data prevalent in this experiment, a characteristic that has been previously discussed in Section~\ref{subsubsec:vrefresults}a.

\subsubsection{Operational Amplifier Circuit}
\label{subsubsec:opampresults}

This section focuses on anomaly detection in the Opamp circuit, illustrated in Figure~\ref{fig:circuits_spice}b. For this analysis, we concentrate on component-level examination since block-level analysis does not facilitate the injection of anomalies into individual circuit components. Similar to the VRef circuit, we conduct three experiments for the Opamp circuit, namely (1) OM Both, (2) OM PFET, and (3) OM NFET, and utilize both transient and DC analyses. The corresponding results are presented in Figure~\ref{fig:table8acc} and Figure~\ref{fig:table8acc_dc}, respectively.

\begin{figure}[t!]
\begin{subfigure}{0.5\columnwidth}
    \centering
    \includegraphics[width=\linewidth]{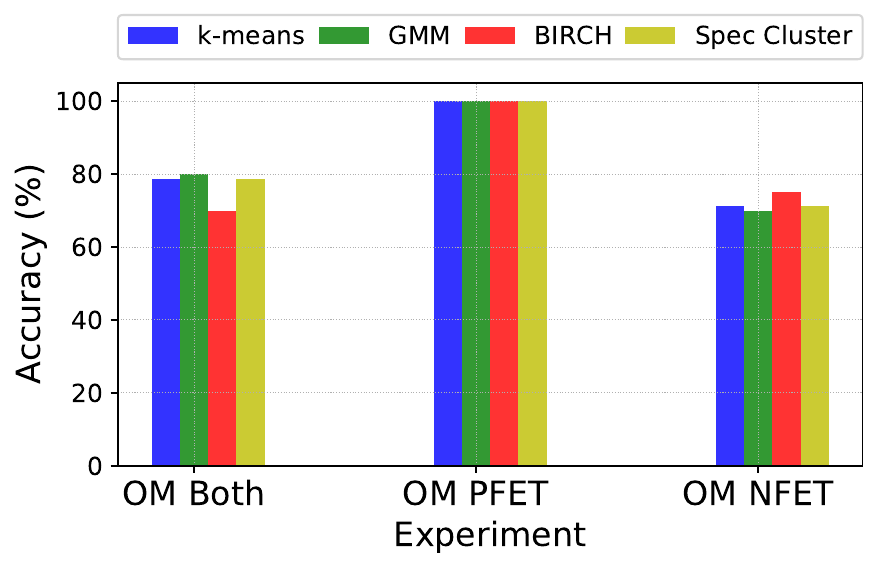}
    \caption{}
    \label{fig:table8acc}
\end{subfigure}%
~
\begin{subfigure}{0.5\columnwidth}
    \centering
    \includegraphics[width=\linewidth]{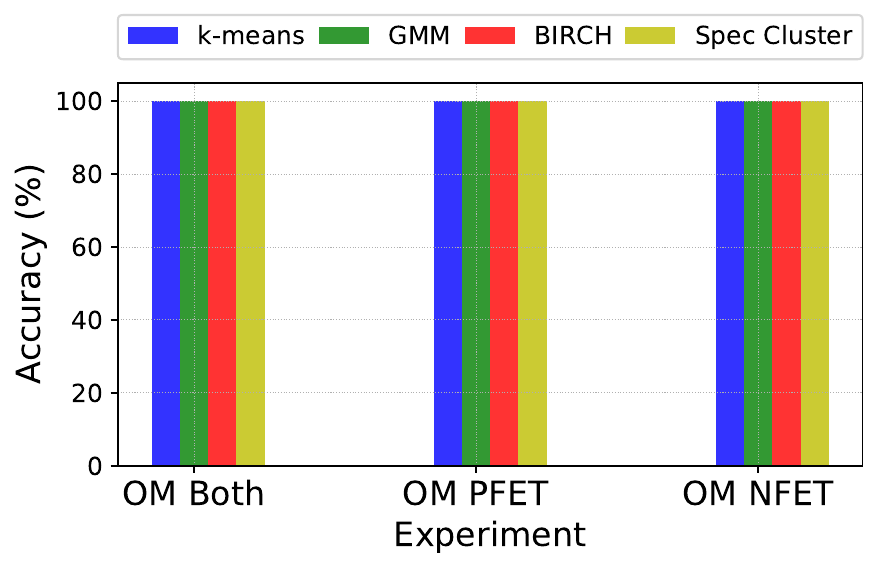}
    \caption{}
    \label{fig:table8acc_dc}
\end{subfigure}
\vspace{0.1in}
\caption{Evaluation of unsupervised learning models for operating mode anomalies pertaining to FETs via (a) transient and (b) DC analysis in operational amplifier circuit.
}
\label{fig:table8results}
\vspace{-0.1in}
\end{figure}

\begin{table}[b!]
\caption{Evaluation of our unsupervised learning-based anomaly detection approach with the inclusion of the proposed centroid selection algorithm.\\
Agg: Aggregated-feature approach; App+A: Framework augmented with Algorithm~\ref{algo:centroidsel_algo}.}
\footnotesize
\centering
\begin{adjustbox}{width=0.9\columnwidth}
\begin{tabular}{|c|c|c|c|}
\hline
\textbf{Best}  & \textbf{Accuracy} & \textbf{App+A} &   \textbf{Accuracy} \\
\textbf{Feature}   & \textbf{(\%)} & \textbf{Accuracy (\%)} &  \textbf{Boost (\%)} \\ \hline \hline

\multicolumn{4}{|c|}{PPA Experiment (VRef circuit)} \\ \hline
Agg & 71 & 74.3 & 3.3 \\ \hline

\multicolumn{4}{|c|}{PRA Experiment (VRef circuit)} \\ \hline
Agg & 52.9 & 56 & 3.1  \\ \hline

\multicolumn{4}{|c|}{Par Fault Experiment (VRef circuit)} \\ \hline
Mean & 75 & 79 & 4 \\ \hline

\multicolumn{4}{|c|}{OM Both Experiment (VRef circuit)} \\ \hline
Mean & 75 & 79 & 4 \\ \hline

\multicolumn{4}{|c|}{OM Both Experiment (Opamp circuit)} \\ \hline
Mean & 79 & 84 & 5  \\ \hline

\end{tabular}
\end{adjustbox}
\label{table:kNNresults_centroidalgo}
\end{table}

As evident from Figure~\ref{fig:table8acc}, GMM achieves the best detection accuracy in the majority of the experiments, with up to 100\% accuracy obtained for the OM PFET experiment by observing the mean of the Opamp output. As previously discussed in Section~\ref{subsubsec:vrefresults}a, the main factor contributing to the superior performance of GMM is the Gaussian distribution of data associated with this series of experiments. 
\textcolor{black}{In contrast to the VRef circuit, transient analysis of the Opamp circuit yields subpar anomaly detection results for the OM NFET experiment. This disparity can be attributed to the relatively reduced influence on the circuit output caused by anomalous NFETs, compared to their counterparts in the VRef circuit.} However, both sets of DC analysis demonstrate 100\% detection accuracy for all experiments (as shown in Figure~\ref{fig:table8acc_dc}), significantly enhancing the anomaly detection performance and underscoring the effectiveness of our proposed solution.

\subsubsection{Centroid Selection Algorithm}
\label{subsubsec:centroidalgoresults}


In this section, we present the results of integrating the proposed centroid selection algorithm into the anomaly detection framework. Table~\ref{table:kNNresults_centroidalgo} provides a comprehensive overview of the results obtained for both the VRef and Opamp circuits. The first column in Table~\ref{table:kNNresults_centroidalgo} specifies the best feature to be observed in each experiment. Columns 2 and 3 present the accuracy of the framework with and without Algorithm~\ref{algo:centroidsel_algo}, respectively. The last column indicates the relative improvement in detection performance achieved by incorporating the centroid selection algorithm.

As discussed in Section~\ref{sec:exp_results}, GMM exhibits the best performance and serves as the base clustering algorithm for augmenting our framework with the proposed centroid selection approach. As shown in Table~\ref{table:kNNresults_centroidalgo}, Algorithm~\ref{algo:centroidsel_algo} improves the anomaly detection performance in all experiments conducted in this section. For instance, in the OM Both experiment of the Opamp circuit, we achieve a promising accuracy of 84\%, representing up to a 5\% improvement in accuracy compared to the previous approach. This enhancement signifies an improvement in detection fidelity and reinforces the effectiveness of the proposed centroid selection algorithm. 

In conclusion, our results demonstrate that the integration of the centroid selection algorithm, to determine the optimal cluster centroids, substantially improves the anomaly detection accuracy, thereby contributing to the overall robustness and reliability of the anomaly detection framework.

\subsubsection{Time Series-based Analysis}
\label{subsubsec:timeseriesresults}

\begin{table}[b!]
\caption{Evaluation of the proposed solution with the inclusion of time series analysis.\\
App+A+TS: Framework enhanced with Algorithm~\ref{algo:centroidsel_algo} and time series approach, WS: Window size.}
\footnotesize
\centering
\begin{adjustbox}{width=0.95\columnwidth}
\begin{tabular}{|c|c|c|c|c|}
\hline
\textbf{Best}  & \textbf{App+A} & \textbf{WS} & \textbf{App+A+TS} & \textbf{Accuracy} \\
\textbf{Feature}  & \textbf{Accuracy (\%)} &    & \textbf{Accuracy (\%)} &  \textbf{Boost (\%)} \\ \hline \hline

\multicolumn{5}{|c|}{PPA Experiment} \\ \hline
Agg & 74.3 & 300 & 80.6 & 6.3 \\ \hline

\multicolumn{5}{|c|}{PRA Experiment} \\ \hline
Agg & 56    & 300 & 60    & 4  \\ \hline

\multicolumn{5}{|c|}{Par Fault Experiment} \\ \hline
Mean & 79   & 750 & 85  & 6 \\ \hline

\multicolumn{5}{|c|}{OM Both Experiment (VRef circuit)} \\ \hline
Mean & 79   & 300 & 83.9  & 4.9 \\ \hline

\multicolumn{5}{|c|}{OM Both Experiment (Opamp circuit)} \\ \hline
Mean & 84   & 300 & 88    & 4  \\ \hline

\end{tabular}
\end{adjustbox}
\label{table:kNNresults_timeseries}
\end{table}

This section presents the results of the time series-based analysis conducted in our experiments. Table~\ref{table:kNNresults_timeseries} provides an overview of the results obtained for both the VRef and Opamp circuits. The first and second columns indicate the best feature to be observed and the accuracy of the framework augmented with Algorithm~\ref{algo:centroidsel_algo}, respectively. Column 3 specifies the optimal window size for this time series analysis, representing the window size that yields the highest accuracy. The fourth column presents the anomaly detection accuracy achieved through the proposed time series-based technique, while column 5 represents the improvement in detection performance compared to the approach without the time series analysis (App+A).

The time series-based analysis enhances the detection performance of the previous approach (App+A) by yielding an improvement in accuracy across all experiments. For instance, in the OM Both experiment of the Opamp circuit, the time series-based approach achieves up to 88\% accuracy by observing the mean of the \textit{Output} signal using a window size of 300 (i.e., $N/5$). Similarly, in the Par Fault experiment for the VRef circuit, the App+A+TS approach achieves a maximum improvement in performance of 6\% (85\% vs. 79\% accuracy) by observing the mean of the \textit{Output} signal using a window size of 750 (i.e., $N/2$). Furthermore, the implementation of the time series approach in other scenarios, such as the TA experiment in the VRef circuit and the OM PFET experiment in the Opamp circuit, also yielded excellent results, achieving 100\% accuracy using a window size of 750 (i.e., $N/2$).

These results demonstrate that the time series-based analysis significantly enhances anomaly detection performance of the proposed solution. By considering temporal patterns and incorporating time series information, \emph{i.e.}, multiple sets of features rather than a single set of features used in the previous analyses (Section~\ref{subsubsec:vrefresults}, Section~\ref{subsubsec:vref_spiceresults}, Section~\ref{subsubsec:opampresults}), the proposed approach achieves improved accuracy and robustness in detecting anomalies in automotive AMS circuits.

\textbf{High Fidelity and Expedited Anomaly Detection:}
Our time series-based approach significantly enhances and expedites anomaly detection performance by capitalizing on the continuous time domain characteristics inherent in AMS circuits. As depicted in Figure~\ref{fig:timeseries_earlyanomdetection}, we incorporate windows of size $N/k$ in our time series-based approach. Utilizing a window size of 300 ($N$=1500, $k$=5), our solution is capable of identifying anomalies as soon as within the initial window, encompassing only 20\% of the signal's time period, while maintaining a commendable detection accuracy of 100\%. For instance, in the case of the Opamp circuit, which operates with Alternating Current power at a clock frequency of \SI{1}{\giga\hertz}, the non-time series technique requires \SI{20}{\micro\second} to detect anomalies. In contrast, the proposed time series-based approach achieves anomaly detection in as little as \SI{4}{\micro\second}. \textbf{This corresponds to a remarkable 5$\times$ reduction in the latency associated with detecting anomalies.}

\subsubsection{Anomaly Abstraction}
\label{subsubsec:anomalyabstractionresults}

\begin{figure}[t!]
\centering
\includegraphics[width=\linewidth]{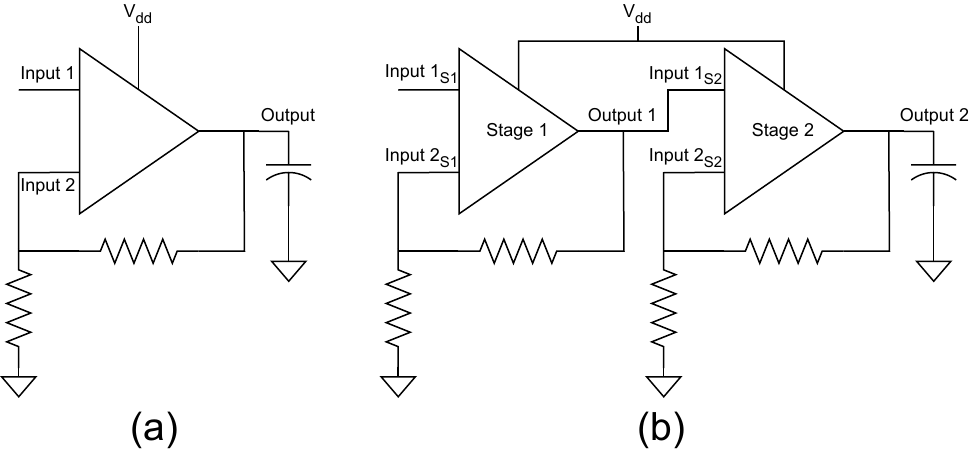}
\caption{Illustration of (a) Single-stage and (b) Dual-stage non-inverting amplifiers.}
\label{fig:kstageamplifiers}
\end{figure}

\textcolor{black}{In this study, we conducted a comprehensive investigation into the performance of four widely used clustering algorithms \textemdash GMM, k-means Clustering, BIRCH, and Spectral Clustering \textemdash in the context of anomaly abstraction. The experiments were performed on three different amplifier stages: Single-stage, Dual-stage, and Tri-stage amplifiers, where each design is built from a base operational amplifier circuit. These amplifier representations have been illustrated in Figure~\ref{fig:kstageamplifiers}(a), Figure~\ref{fig:kstageamplifiers}(b), and Figure~\ref{fig:threestageamp}, respectively. Specifically, we introduce anomalies in a k-stage amplifier by simulating an anomalous operation of the base opamp symbol, i.e., OM Both (described in Section~\ref{subsubsec:opampresults}), at various locations in the amplifier. The algorithms are evaluated in terms of the anomaly detection accuracy furnished in the various experiments.}

\begin{figure}[t!]
\centering
\includegraphics[width=0.8\linewidth]{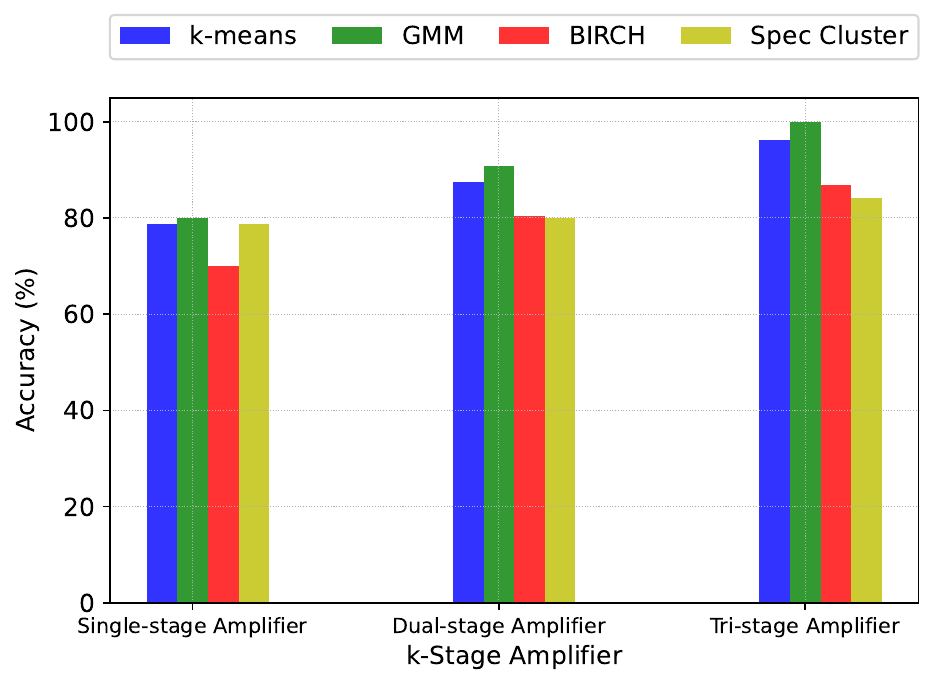}
\caption{Performance of various clustering algorithms in the context of k-stage inverting amplifier.}
\label{fig:kstageamplifier_results}
\end{figure}

\textcolor{black}{As evident from Figure~\ref{fig:kstageamplifier_results}, the results reveal valuable insights into the efficacy of the examined algorithms across the various amplifier configurations. Notably, the GMM algorithm consistently demonstrated superior detection performance across all three amplifier stages, achieving 80\% accuracy in the Single-stage amplifier, 90.8\% in the Dual-stage amplifier, as well as an ideal performance of 100\% in the Tri-stage amplifier. Similarly, the k-means clustering algorithm exhibited competitive performance, yielding 78.75\%, 87.5\%, and 96.2\% detection accuracy in the respective amplifier stages. BIRCH and Spectral Clustering, while also displaying promising results, presented a slightly lower performance compared to GMM and k-means, with BIRCH achieving 70\%, 80.5\%, and 86.8\% accuracy, and Spectral Clustering reaching 78.75\%, 80\%, and 84.2\% accuracy, respectively.}

\textcolor{black}{Overall, these findings provide valuable guidance to designers and researchers seeking to optimize amplifier performance. The study showcases the GMM and k-means Clustering algorithms as particularly promising choices for detecting anomalies in diverse amplifier configurations, offering superior detection performance. \textcolor{black}{By analyzing the strengths and weaknesses of each algorithm in distinct amplifier stages, this research can contribute to the advancement of clustering-based algorithms in the realm of amplifier design and may pave the way for more reliable and robust amplifier circuits in various applications or use cases.}}

\subsubsection{Algorithm Analysis}
\label{subsubsec:algorithmanalysis}


Based on our experimental results, it is evident that GMM outperforms the other clustering algorithms, namely k-means, BIRCH, and Spectral Clustering, in that specific order. The superior performance of GMM can be attributed to the nature of the datasets generated from the circuits. Since a significant portion of our experiments exhibits a Gaussian distribution of data, GMM is adept at capturing and modeling such data distributions, leading to superior detection performance among the clustering algorithms evaluated. As a result of these findings, we adopt GMMs as the primary clustering algorithm in our anomaly detection framework. By leveraging the strengths of GMMs, we can effectively detect anomalies and ensure reliable and accurate performance in detecting various types of faults and anomalies in automotive AMS circuits.

\section{Discussion: Frequently Asked Questions}
\label{sec:discussion}

\balance

\noindent
\textcolor{black}{\textbf{Q1:} What is the duration required for anomaly detection?}

\noindent
\textcolor{black}{\textit{Ans:} 
Anomaly detection time is influenced by resource availability, such as power and bandwidth budgets. The versatility of the proposed framework allows for its deployment on various platforms, including machine learning accelerators, general-purpose CPUs and GPUs, as well as cloud-based platforms~\cite{su2018improving}. Our experimental setup, which involves a system equipped with an Intel i7 8th generation processor and 16 GB RAM, yielded an anomaly detection time of \SI{15}{\milli\second}.
}
\smallskip

\noindent
\textcolor{black}{\textbf{Q2:} How can we ascertain if early anomaly detection precedes the occurrence of the failure state?}

\noindent
\textcolor{black}{\textit{Ans:} 
In the context of AMS circuits in automotive systems, anomalies or errors typically undergo a period of propagation from the circuit level to the system level, \emph{i.e.}, from the hardware to the application hosted by the system. Additionally, such fault or error manifestation at the application layer involves an additional time delay. Consequently, within this error propagation and manifestation time period, an opportunity arises for early anomaly detection before an actual system failure occurs. This early detection window allows for the implementation of proactive error and fault mitigation strategies, thereby potentially averting system failures and improving overall system reliability.
}

Prior approaches have integrated semi-supervised machine learning techniques to enable early anomaly detection in automotive AMS circuits, wherein anomaly scores are utilized to characterize the anomalous behavior of signals~\cite{su2018improving}. These scores serve as an evaluation metric quantifying the deviation from normal profiles and can potentially be associated with a test specification limit, representing a user-defined threshold beyond which a test would be considered a failure. This correlation enables the classification of different ranges of anomaly scores, such as, a high anomaly score corresponding to a test specification violation, a low anomaly score indicating a pass case with high confidence, and a medium anomaly score falling between these two extremes. Consequently, this offers an opportunity to identify anomalies before test specification violations, thereby facilitating early anomaly detection.

\smallskip

\noindent
\textcolor{black}{\textbf{Q3:} How does our solution approach measure up when compared to existing techniques?}

\noindent
\textit{Ans:} 
The principal distinction between the proposed approach and existing strategies~\cite{su2018improving} resides in the anomaly detection framework. In~\cite{su2018improving}, a semi-supervised learning strategy is employed, wherein ML algorithms are trained using normal, non-anomalous data to predict anomalies in the test data collected during mission mode. Conversely, our framework entails the construction of a comprehensive training dataset that encompasses diverse anomalous scenarios. Subsequently, an unsupervised learning algorithm is utilized to generate clusters of signals, distinguishing non-anomalous from anomalous signals, both during training and inference phases. In contrast to~\cite{su2018improving}, our proposed method incorporates feature extraction as an integral component of the anomaly detection model, rendering it suitable for handling large feature spaces.

\smallskip

\noindent
\textcolor{black}{\textbf{Q4:} Why are bandgap voltage reference and operational amplifier circuits considered?}

\noindent
\textit{Ans:}
To enhance the relevance of the experimental setup in alignment with contemporary scope and state-of-the-art systems, the present study assesses our proposed solution through a case study involving two AMS circuits commonly integrated into contemporary automotive SoCs. These circuits include the bandgap voltage reference circuit and operational amplifier circuit. \textcolor{black}{Moreover, as a part of our prospective endeavors, we aim to extend our investigation to higher abstraction levels, such as the application level, hosted by the system. Additionally, we plan to extend our investigations to encompass other analog and mixed-signal circuits found within automotive system-on-chips. For instance, voltage regulators, analog-to-digital converters, digital-to-analog converters, and other related AMS circuit representations.}

\section{Conclusion}
\label{sec:conclusion}


This research paper presents a novel framework for early anomaly detection in AMS components of automotive SoCs, with the primary objective being the enhancement of their FuSa. The utilization of unsupervised machine learning is motivated by its flexibility during training and its capability to discover latent patterns within datasets. Unlike supervised learning, unsupervised learning does not necessitate well-labeled datasets for training and evaluation, which is particularly advantageous when dealing with automotive systems, where such datasets are scarce and challenging to generate. This paper addresses several challenges encountered by automotive AMS components, which have a direct impact on their functional safety. Firstly, an anomaly dataset is constructed by systematically injecting anomalies at various locations within the circuit, accounting for different anomalous scenarios. This comprehensive dataset enables thorough evaluation and validation of the proposed framework. Secondly, a novel centroid selection algorithm is introduced to identify optimal cluster centroids. This algorithm enhances the fidelity of anomaly detection by effectively capturing the characteristics of normal and anomalous data points, facilitating accurate classification. Thirdly, the paper employs time series-based analysis techniques to further improve the performance of anomaly detection. This approach captures temporal patterns and dependencies, thereby furnishing a more effective detection performance and reduced latency compared to non-time-series-based approaches. We evaluate the effectiveness of the proposed framework using two AMS circuits (bandgap voltage reference circuit and operational amplifier circuit) that are representative of modern automotive systems. The results demonstrate up to 100\% accuracy and a five-fold (5$\times$) reduction in detection latency compared to non-time-series-based approaches. \textcolor{black}{Furthermore, we performed anomaly abstraction to evaluate the impact of anomalies as they propagate from a lower abstraction level (component-level) to a higher abstraction level (block-level). Specifically, we considered a k-stage amplifier design, wherein we constructed single-stage, dual-stage, and tri-stage non-inverting amplifiers using a base opamp circuit. The proposed anomaly detection framework furnishes up to 100\% detection accuracy in the tri-stage amplifier.} These findings validate the efficacy of the proposed solution in improving the functional safety of automotive AMS circuits. \textcolor{black}{As part of future work, our research aims to explore higher abstraction levels, such as the application-level, and larger AMS circuits. This would further enhance the effectiveness of the proposed approach, advocating its use as a holistic SoC-level automotive AMS FuSa violation detection solution.}


\bibliographystyle{IEEEtran}
\bibliography{references.bib}

\newpage

\end{document}